# CoGraM: Context-sensitive granular optimization method with rollback for robust model fusion

**Author:** Julius Lenz

## Abstract

Merging neural networks without retraining is central to federated and distributed learning. Common methods such as weight averaging or Fisher merging often lose accuracy and are unstable across seeds. CoGraM (Contextual Granular Merging) is a multi-stage, context-sensitive, loss-based, and iterative optimization method across layers, neurons, and weight levels that aligns decisions with loss differences and thresholds and prevents harmful updates through rollback. CoGraM is an optimization method that addresses the weaknesses of methods such as Fisher and can significantly improve the merged network.

## Introduction

Federal and distributed learning requires methods that combine multiple models without central re-training (1). A good example of this would be the combination of neural networks from different hospitals, each with slightly different training data (e.g., image resolution, different patient groups) (1,3,4). Classic approaches such as FedAvg (2) and Fisher Merging (5) are more prone to accuracy losses and instability when dealing with heterogeneous data distributions (5–7). CoGraM addresses and improves this through three core features:

(1) **Adaptive granularity:** Decisions are made adaptively at the layer, neuron, and weight levels. Granularity is refined as needed (Figure 1).

(2) **Context Sensitivity:** Evaluation decisions are based on loss differences (Equation 2), measured on representative batches or prototypes (see prototype calculation). For each decision, the context of the respective parameter is taken into account. This means that the value of a parameter is always evaluated in its current position within the neural network $M$ and with respect to its impact on loss reduction. The exact procedure is described in the Methods section.

(3) **Rollback:** Changes or merges at a finer level of granularity are only adopted if they lead to improvement (Equation 7); otherwise, they are locally reverted to ensure the best possible outcome.

The goal of CoGraM is to improve the accuracy of a base network. The procedure is carried out step by step with adaptively refined granularity (Layer → Neuron → Weight) (Figure 1), where the granularity is increased for finer merging when necessary. Changes are only accepted if they demonstrably reduce the loss $L$ (Equation 2) of the fused network $M$; otherwise, a rollback is performed. Decisions in CoGraM are fully context-sensitive, meaning that model parameters are evaluated based on the current state of network $M$. In this way, a consistent model adapted to the target data is created even without fine-tuning.

In contrast to other merging approaches that treat all parameters equally (8,9), CoGraM evaluates each structure (layer, neuron, weight) individually with respect to its contribution to loss reduction. Only demonstrable improvements are adopted - otherwise, rollback is applied.

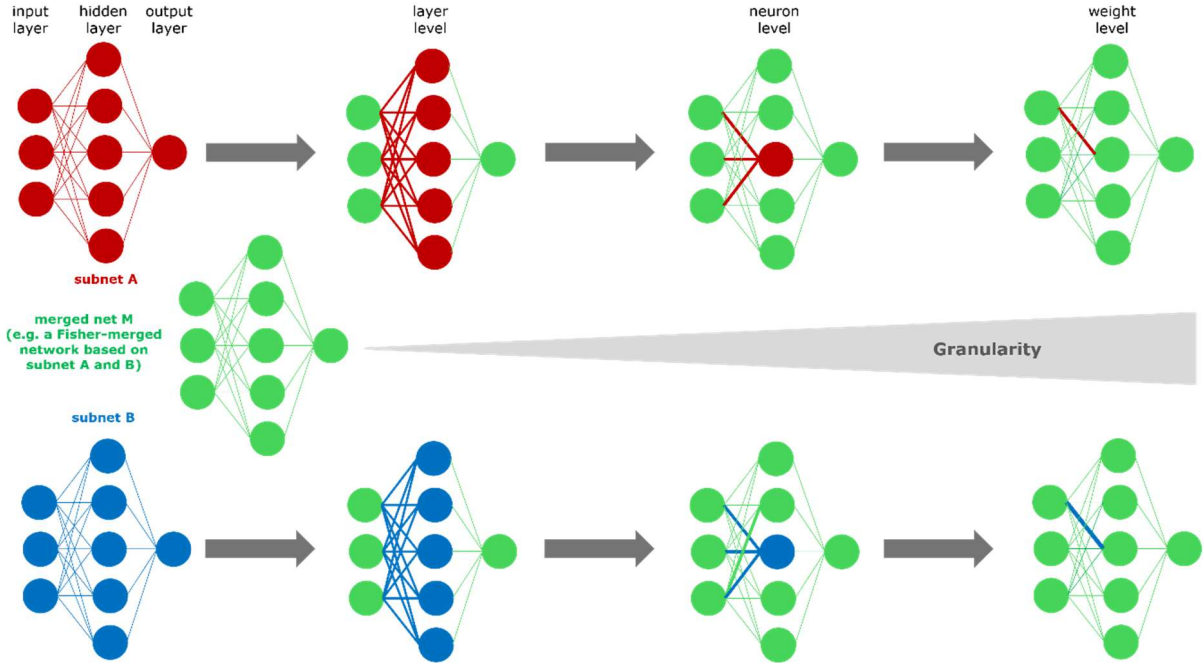

*Figure 1: Schematic representation of granularity refinement.*

## Methods

### Evaluation Data and Loss

**Data Basis:** Decisions are made on a stable evaluation foundation, namely the loss $L$ (Equation 2) with respect to the combined training dataset of subnetworks A and B. These decision processes rely on the loss of the network $M$ currently undergoing the merge process, which is calculated using various prototypes or batches. Parts of subnetworks A and B are inserted into $M$, and the loss of $M$ is computed so that both candidate states can be compared.

**Batches:** One or more random batches are drawn from the combined training dataset of A and B. Depending on the desired accuracy, the size of these batches can be varied or set equal to the size of the entire training dataset.

**Prototypes:** Within the batches, training samples are clustered according to their targets (the objectives on which the subnetworks are trained—for a classifier, these correspond to the individual classes), e.g., using one-hot encoding or k-means clustering (10,11). Representatives for each cluster (based on the training target) are then formed using the geometric mean (12). The advantage of the geometric mean is increased robustness to deviations and a stronger emphasis on commonalities in the data (12,13). The goal is to reduce both the variance of the target data and runtime, since fewer comparison samples (prototypes) mean fewer evaluations are required to determine the value of a structure (layer, neuron, weight).

*Equation 1*: Formation of a class prototype $\tilde{x}_c$ as the geometric mean of the magnitude vectors $|x|$ of all samples $x \in S_c$, stabilized by $\varepsilon$ and normalized by the class size $|S_c|$.

$$\tilde{x}_c = \left( \prod_{x \in S_c} (|x| + \varepsilon) \right)^{\frac{1}{|S_c|}}$$

**Loss Measurement:** A cross-entropy loss function (14) evaluates the candidate states of subnetworks A and B within the fused network $M$ with respect to the generated prototypes $\tilde{x}_c$ (Equation 1). It is important to note that a cross-entropy loss function is not always appropriate; for a regression network, for example, it should be replaced by mean squared error (MSE) (15).

The calculation is always performed at the respective level of granularity (layer, neuron, or weight) and proceeds backward through network $M$. The last layer is processed first, and the first layer last, following the principle of backpropagation (16). All comparisons between the candidate states of subnetworks A and B use identical evaluation data to ensure consistency and comparability. In each case, the states of subnetworks A and B are inserted into network $M$. Depending on the granularity, such a state may correspond to a layer, a neuron, or a weight.

In the current CoGraM variant for classification networks, the softmax function (17) is additionally applied to transform the raw network outputs (logits) into probability distributions over the classes. This allows the cross-entropy loss $L_n$ (Equation 2) to precisely determine how well the model assigns the prototypes $\tilde{x}_c$ to the target classes $\hat{y}_c$. The closer the softmax probabilities (17) are to the target classes, the lower the resulting loss $L_n$.

*Equation 2: Computation of the mean cross-entropy loss $L_n$ for subnetwork n, based on the prototypes $\tilde{x}_c$ and the target distributions $\hat{y}_c$. The evaluation measures how well the model with parameters W captures the class representation via the softmax $\sigma(f(\tilde{x}_c;W))$. The superscript T in $\hat{y}_c^T$ denotes the transposition of the target vector, enabling the formation of a scalar product with the log-softmax vector.*

$$L_n = -\frac{1}{C}\sum_{c=1}^{C} \hat{y}_c^T \log \sigma\left(f(\tilde{x}_c;W)\right)$$

**Evaluation Procedure:** Within each level of granularity, all relevant structures are systematically examined. Following the principle of backpropagation (10), all layers of the network are processed from back to front at the layer level. At the neuron level, every neuron within the currently considered layer is individually evaluated. Finally, at the weight level, the individual weights of each neuron are inspected. This ensures that every structure is assessed independently and that no level is omitted.

## Merging and Weighting

**Mixing Factor α:** A mixing factor $\alpha$ between 0 and 1 controls the combination of the candidate states from subnetworks A and B. $\alpha$ is derived from the loss difference using the sigmoid function (Equation 4). Larger advantages in loss reduction therefore lead to a stronger weighting of the candidate states during fusion.

**Adoption Principle:** Only those parameters are adopted into the combined network that result in an improvement with respect to the network's loss, as they are weighted according to the mixing factor $\alpha$.

**Loss Difference ΔL:** For each evaluated structure, the loss $L$ of the fused network $M$ is calculated (across all prototypes). This is done both with the parameters inserted from subnetwork A and with those from subnetwork B. The difference between these loss values, $\Delta L$ (Equation 3), forms the basis for deciding whether to merge directly or to refine further.

*Equation 3: Computation of the loss difference ΔL between subnetworks A and B, where $L_A$ and $L_B$ represent the loss values of the respective candidate states of subnetworks A and B.*

$$\Delta L = L_A - L_B$$

**Thresholds τ:** For each level of granularity, defined thresholds $\tau$ exist. These include the minimum improvement threshold $\tau_{min}$ and the upper granularity threshold $\tau_{max}$. Decisions regarding refinement are guided by the loss difference $\Delta L$, $\tau_{min}$, and $\tau_{max}$:

- **Case 1:** If $|\Delta L| < \tau_{min}$, the decision is considered uncertain, and the process moves to the next finer level in order to refine the decision locally.

- **Case 2:** If $|\Delta L| > \tau_{max}$, the decision is considered too coarse, and refinement is carried out at the next finer level.

- **Case 3:** If $\tau_{min} \leq |\Delta L| \leq \tau_{max}$, then $|\Delta L|$ lies within the acceptable range, and merging is performed directly at the current level (Equation 6).

For Cases 1 and 2, the rollback mechanism described earlier is applied. Thresholds $\tau$ can be defined separately for each level (layer, neuron, weight) and adapted to the specific application. If the loss difference $\Delta L$ between candidate states A and B at the current granularity level does not fall within the acceptable range of Case 3, the decision is classified as uncertain or insufficient. In this case, refinement is performed according to Equations 4 and 5.

*Equation 4: Computation of the mixing factor α using the sigmoid function (Equation 5), based on the loss L of the candidate states A ($L_A$) and B ($L_B$), and the steepness parameter λ, which controls the hardness of the mixture (i.e., the tendency toward the better subnetwork). According to Equation 3, ΔL is the difference between $L_A$ and $L_B$. The mixing factor α is recalculated in the current context each time two candidate states are fused.*

$$\alpha = \frac{1}{1 + e^{\lambda(L_A - L_B)}} = \frac{1}{1 + e^{\lambda \cdot \Delta L}}$$

This $\alpha$ is calculated, depending on the current granularity level, either for each layer, each neuron, or even for every single weight.

*Equation 5: Sigmoid-Function*

$$\sigma(x) = \frac{1}{1 + e^{-x}}$$

*Equation 6: Convex linear combination for mixing the candidate states A and B with the mixing factor α (Equation 4). Here, $W_A$ and $W_B$ denote the weights of the subnetworks. At the layer level, this corresponds to a weight within the layer; at the neuron level, it is a weight within the neuron. $W_{new}$ is the resulting weight that is inserted into M.*

$$W_{new} = \alpha \cdot W_A + (1 - \alpha) \cdot W_B$$

**Layer Level**

- **Evaluation:** The layer parameters from models A and B are compared using loss measurement. All layers of $M$ are processed from back to front, with each layer of $M$ replaced by the corresponding layer from A or B. For every inserted layer from A or B, the loss of $M$ is calculated. This allows determination of which part from A or B contributes more strongly to loss reduction.
- **Decision:** If $\Delta L$ lies within the acceptable range (Case 3), the two inserted layers from subnetworks A and B are convexly combined using the mixing factor $\alpha$. At the layer granularity, $\alpha$ is calculated once for the entire layer. All layer parameters are then combined through a convex linear combination (Equation 6). This is performed directly on network $M$, with the fused part of the subnetworks replacing the corresponding part in $M$. If $\Delta L$ does not fall within the acceptable range (Case 1 or Case 2), refinement proceeds to the individual neurons within the examined layer. This principle is illustrated schematically in Figure 2.
- **Note:** At the layer level, rollback does not exist. Here, only an increase in granularity can be applied if necessary.

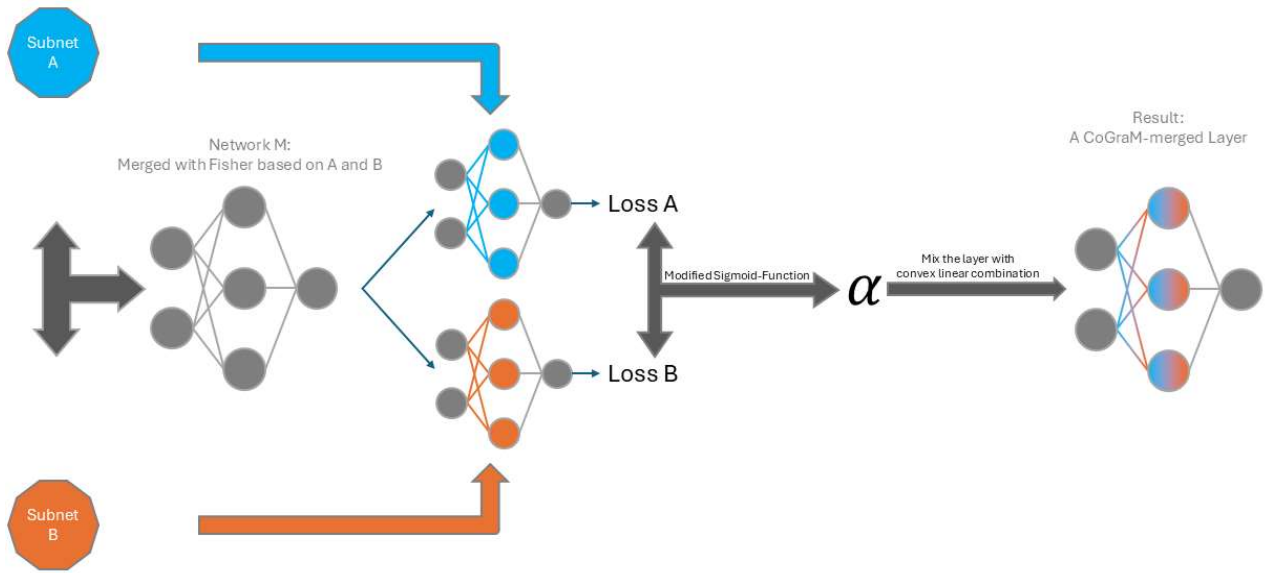

*Figure 2: Schematic representation of the mixing process of a layer according to CoGraM. The mixing process at the neuron or weight level follows the same principle.*

**Neuron Level**

- **Evaluation:** For each neuron of the layer being refined, the corresponding neuron from subnetwork A and from subnetwork B is inserted into $M$. In both cases, the loss of $M$ is calculated, allowing comparison of which candidate contributes more strongly to loss reduction.
- **Decision with Rollback:** If $\Delta L$ of the tested neurons from subnetworks A and B is within the acceptable range (Case 3), these neurons are combined using the mixing factor $\alpha$ and the convex linear combination (Equation 6). At the neuron granularity, $\alpha$ is calculated once per neuron. All neuron parameters are then combined through a convex linear combination. Changes are only adopted if they reduce the loss of *M* compared to the previous state: if a fused neuron does not improve the loss relative to the same neuron fused at the layer level, then local rollback restores that neuron to the parameters from the prior layer-level fusion. It should be noted that rollback is local, meaning that some neurons may revert to the layer-level state while others remain fused. In other words, only those neurons that fail to improve the loss of *M* compared to their layer-level parameters are rolled back.
- **Refinement:** If the loss difference at the neuron level does not fall within the acceptable range, refinement proceeds further to the weight level. Otherwise, the neurons from A and B are weighted by $\alpha$ and fused into $M$.

**Weight Level**

- **Evaluation:** The individual incoming weights of a neuron are examined. As already described for layers and neurons, the weights from subnetworks A and B are each inserted into the corresponding position and compared through loss evaluation.
- **Decision with Rollback:** Since no finer granularity follows after the weight level, the weights are merged and then adopted (Equation 6) if an improvement compared to the previous loss at the neuron level can be achieved. At the weight level, the mixing factor $\alpha$ is calculated separately for each weight. The weights are then combined again using a convex linear combination (Equation 6). If no improvement occurs, the respective weight is replaced by the weight that was fused at the higher neuron level (local rollback).
- **Finalization:** After local optimization, all accepted changes are consolidated into the fused model state.

*Equation 7: Rollback procedure – the new loss $L_{post}$ is only accepted if it is smaller than the previous loss $L_{pre}$.*

$$accept\ changes = \begin{cases} true, if\ L_{post} < L_{pre} \\ false, if\ L_{post} \geq L_{pre} \end{cases}$$

## Iterative Procedure

The described CoGraM methodology is typically applied iteratively: initially, *M* is initialized as an already merged network (e.g., using the Fisher merge method (5)). The first CoGraM iteration is then performed on $M$, and CoGraM can subsequently be repeated any number of times—always operating on the most recently fused network $M$. This network then serves as the initialization for the subsequent CoGraM iteration (Equation 8).

*Equation 8: Formal representation of the iterative procedure with the fused network M and CoGraM applied to subnetworks A and B.*

$$M^{(k+1)} = CoGraM(M^{(k)}, A, B)$$

### Gradient-Kickoff

CoGraM can be used with a gradient kickoff (18), which increases its sensitivity to subsequent fine-tuning. Before the actual fine-tuning, a short training phase is carried out, referred to as the gradient kickoff. In this phase, the fused model is trained for a few epochs with an elevated learning rate and, in the case of SGD, with momentum (14,19). Typically, Adam (20) is used as the optimizer (18).

1. Perform CoGraM based on a Fisher-merged network $M$.
2. Choose the optimizer (Adam or SGD with momentum).
3. Carry out a small number of gradient descent steps (<10 epochs) on the training data.
4. Use a learning rate significantly higher than that applied in fine-tuning (2–3 times higher).
5. Optionally apply gradient clipping (21) to improve numerical stability.

- **Momentum:** When using SGD, a momentum term $\mu$ (typically 0.9) is employed to account for the direction of previous updates and to smooth the steps.
- **Transition:** After completing the kickoff phase, the learning rate is reduced and regular fine-tuning begins.
- **Goal:** The objective is to make CoGraM more responsive and better prepared for the subsequent fine-tuning.

### The Fused Network M

The resulting model is a fused network $M$ that has only adopted demonstrable improvements and has automatically protected itself against harmful updates through rollback. The iterative approach enables the subnetworks to be combined based on the loss contributions of their individual components. Prototypes assist in this process by reducing the variance of the target dataset, representing it as accurately as possible, and thereby enabling robust optimization.

# Results

The following tests were conducted on classification data, where both subnetworks A and B were expected to learn the properties and characteristics of classes ($n = 20$) and correctly recognize them.

All data were synthetically generated in a 32-dimensional space. Each of the 20 classes was assigned a random center, which was further differentiated by subclusters with random directional shifts to increase variance. For the subclusters, samples were drawn from multivariate normal distributions and subsequently nonlinearly distorted using sine and tangent transformations as well as additive noise (22–25).

Test data were generated separately with different properties, including slightly increased noise (26,27), in order to adequately evaluate the generalization capability of the fused model.

The exact parameter settings and the algorithm used are provided at the end of the Results section.

All subsequent figures show performance only on a few seeds. Larger tests and benchmarks are planned for future work. However, all figures presented here are representative of large-scale tests. For clarity, the figures are kept in a compact format.

## Test 1 - Use of Homogeneous Data

First, the accuracy values of Fisher with and without subsequent CoGraM were compared. Figure 3 shows the results across 30 seeds. Subnetwork data were generated mathematically identically but on a different seed (homogeneous data). Test data were generated on yet another seed with slightly increased noise to evaluate the fused model.

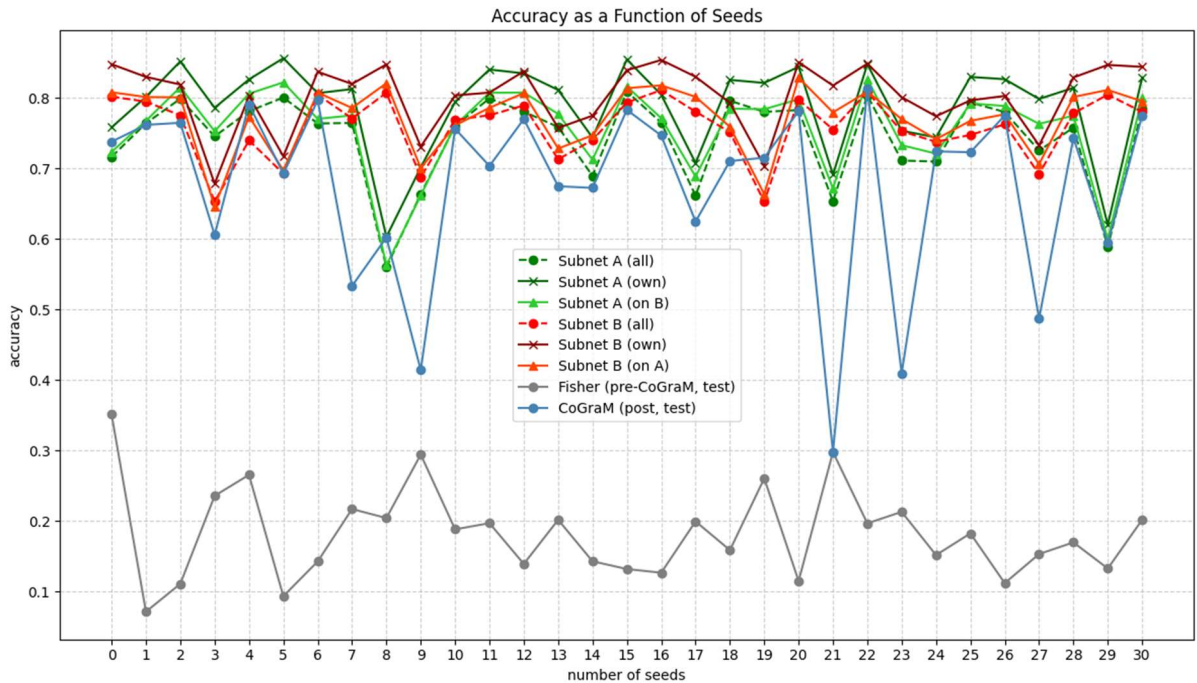

*Figure 3: Accuracy of M with Fisher, with and without subsequent CoGraM procedure. The accuracies of subnetworks A and B were generated across all data constellations. CoGraM was executed solely at the layer granularity level, without fine-tuning or kickoff. For prototype formation, one-hot clustering was used. $\lambda = 5.5$*

It can be observed that Fisher alone shows a considerably weaker average accuracy compared to Fisher combined with CoGraM. However, the graph depicting accuracy after applying CoGraM appears to have a higher standard deviation. In most seeds, this graph is only slightly weaker than the subnetworks themselves, but there are also significant drops, often on seeds where the subnetworks themselves perform poorly. A particularly strong drop can be seen on seed 21, where CoGraM did not improve network *M* at all.

## Test 2 – Use of Heterogeneous Data

Here, the accuracy values of Fisher with and without subsequent CoGraM were compared. Figure 4 shows the results across 30 seeds. The subnetwork data were generated mathematically differently, each on a separate seed (heterogeneous data). Test data were generated on yet another seed with slightly increased noise.

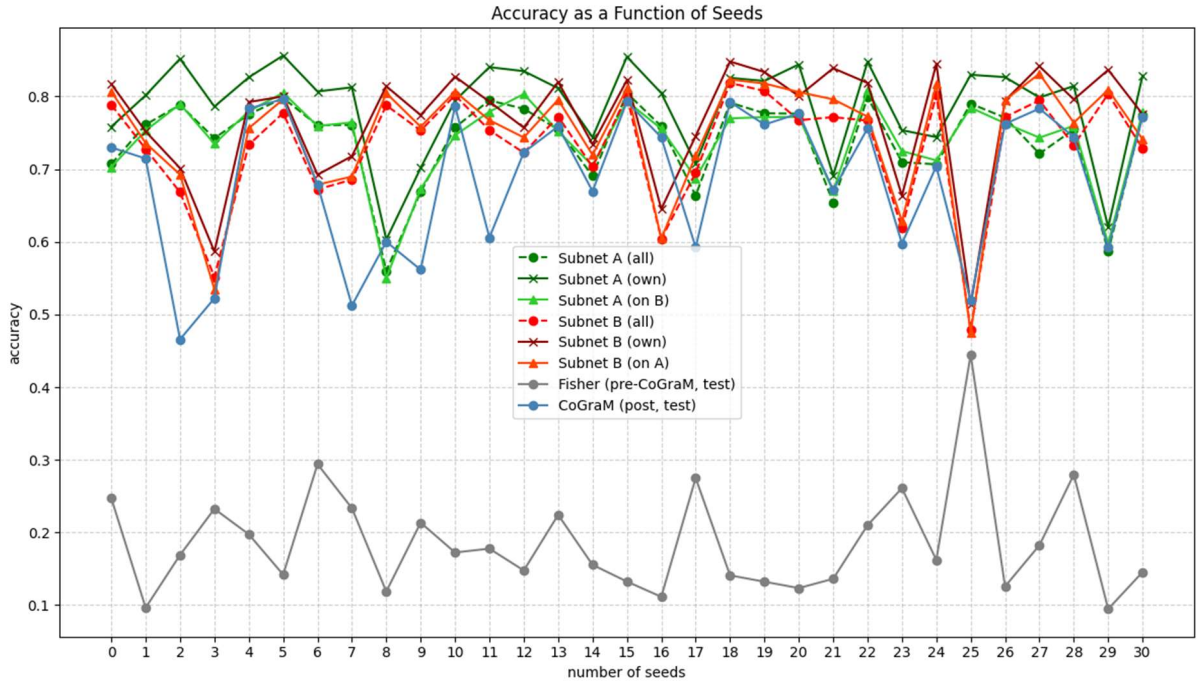

*Figure 4: Accuracy of M with Fisher followed by CoGraM and without subsequent CoGraM. In addition, the accuracies of subnetworks A and B are reported across all data constellations. In this test, CoGraM was executed only at the layer granularity level, without fine-tuning or kickoff. For prototype formation, one-hot clustering was used. $\lambda = 5.5$*

It can also be observed here that CoGraM is clearly superior on average. The severity of the drops decreases in comparison. CoGraM consistently provides an improvement, but on seeds where no drops are observed, it deviates more strongly from the accuracy values of the subnetworks. Overall, the standard deviation also appears to be quite high in this case.

Fusions down to the neuron and weight level were also tested. However, these resulted in higher variances and, on average, lower accuracy. Therefore, the experiments reported here are restricted to the layer level, which proved to be the most stable and effective variant. Further transparent tests will follow in subsequent extended versions.

# Discussion

CoGraM is a novel context-sensitive, granular adaptive optimization/merging method with local rollback. Decisions are based on clearly defined loss differences and thresholds. Refinements are only performed when coarse comparisons are insufficient. Initial tests demonstrate a fusion with relatively few losses and good efficiency. CoGraM thus represents a practical method for model fusion in federated and distributed learning, enabling two or more subnetworks to be combined. However, CoGraM requires an initial network *M* on which the first loss evaluations can be carried out. Fisher merging initially proved to be the most effective and runtime-efficient approach.

Through prototype selection or by setting a granularity limit (e.g., restricting to the layer level), CoGraM can be scaled to larger models in terms of runtime. Nevertheless, the question arises whether a geometric mean per target class is always sufficient when classes are defined by many complex features. In such cases, multiple averages per target cluster might be more appropriate. A principle based on k-means clustering and prototype selection using medoid vectors (28,29) is also conceivable and may be better suited in certain applications than direct averaging, since averaging can obscure important details.

One challenge of CoGraM is the high standard deviation of accuracy across seeds when subnetworks exhibit drops in performance on certain seeds. This often also affects CoGraM. In such cases, CoGraM produces fluctuating results, particularly when granularity extends to the neuron or weight level and the subnetworks show large accuracy differences. A merge restricted to the layer level provides greater stability and higher accuracy in many cases. A pure layer-level merge, without refinement to deeper granularity, currently yields the best results.

We hypothesize that the increasing fluctuations and, on average, lower accuracy values observed when refining granularity are due to the fact that at finer levels many parameters are fused individually, thereby losing their relation to other parameters in the network. For a neural network, it is crucial that parameters remain consistent and aligned so that the network can generalize well and deliver stable results. This consistency may be lost in overly fine merges, leading to strong fluctuations on certain seeds. Neuron- and weight-level merges, while very fine-grained, tend to vary across seeds because the process is highly sensitive and sometimes poorly aligned.

Networks fused with CoGraM (and an initial method) are less sensitive to fine-tuning compared to networks fused only with Fisher. This is likely because Fisher is gradient-based, whereas CoGraM is loss-based. As a result, the Fisher network is already positioned more favorably in the loss landscape, while CoGraM, due to its loss-based approach, must first overcome multiple local maxima or plateaus during fine-tuning.

To address this issue, we introduced gradient kickoff as an additional feature in CoGraM. This makes CoGraM more sensitive and better suited for subsequent fine-tuning. After 8 epochs of kickoff and 20 epochs of fine-tuning, the fused network achieves accuracy clearly above that of the subnetworks. This improvement was demonstrated both on the individual subnetwork data and on the combined dataset. The kickoff, through its elevated learning rate, enables CoGraM to overcome maxima and escape plateaus within a few steps. Consequently, the likelihood increases that CoGraM, after the kickoff, is positioned near a local minimum, allowing subsequent fine-tuning to guide it effectively into a minimum.

### Idea of the Iterative Approach

An iterative CoGraM methodology appears reasonable, since the evaluation of candidate states is loss-based. Once CoGraM has already completed an iteration, the result is a neural network adapted to the target data—i.e., the merged training dataset from A and B—based on a mixture of the subnetworks A and B.

On the basis of an already aligned and consistent network, the evaluation of the loss for candidate states A and B can be performed with greater precision. With each iteration, the most recently fused network serves as the initialization for the subsequent iteration, making the loss evaluation increasingly accurate and consistent.

In this way, the overall accuracy of network $M$ is further improved, as CoGraM itself makes the loss evaluation more consistent through its intermediate results. Even minor fluctuations can be smoothed out within this process.

### Distinction from Existing Approaches

**Granularity Control:** While classical methods treat all model parameters globally (e.g., through averaging), CoGraM makes decisions at the structural level- starting at the layer level and, if necessary, refining down to neurons and weights. This enables local optimization of model fusion.

**Context Sensitivity:** CoGraM does not evaluate candidate states in isolation but in the context of the current model $M$. Loss evaluation is always performed on the fused network, ensuring that the decision for A or B is dynamic and data-driven.

**Iterative Approach:** CoGraM can be applied multiple times. Each iteration uses the result of the previous network $M$ as its starting point, making loss evaluation increasingly consistent. Comparable iterative self-refinement is not foreseen in classical methods.

**Rollback Mechanism:** Unlike methods such as Fisher, which adopt changes directly, CoGraM checks each modification for actual loss improvement. If no improvement is achieved, a local rollback is performed—serving as a safeguard to maintain the quality of the fused model.

**Gradient Kickoff:** Since CoGraM is not gradient-based but relies on loss comparisons, the model is set into motion after fusion through a short kickoff phase. This improves the starting position for subsequent fine-tuning—a step not included in methods such as FedAvg or Fisher.

### Conclusion

CoGraM is a novel, context-sensitive, loss-based optimization method for AI models and neural networks. As an optimizer applied to an already fused network, CoGraM achieves significantly higher accuracy. It distinguishes itself as a powerful optimization approach, though it is partly susceptible to bias toward the better subnetwork when the subnetworks exhibit strongly differing accuracies.

## Outlook

The central goal is to equip CoGraM with a regularization mechanism so that subnetworks with large loss differences can be better combined. Such an extension would address one of the current weaknesses of a loss-based method and significantly improve CoGraM, particularly in extreme cases.

In addition, work is underway to enable better fusion with lower standard deviation across seeds, as well as stronger generalization and accuracy at the neuron and weight levels. A particularly promising concept would be one that measures the information flow through the network and detects disturbances or inconsistencies in the parameters.

Further efforts will focus on generating additional results, benchmarks, and comparisons. This work is a preprint intended to introduce the idea of CoGraM.